\documentclass{article}

\PassOptionsToPackage{numbers, compress}{natbib}

\usepackage[preprint]{neurips_2022}

\usepackage{graphicx}
\usepackage{caption}
\usepackage{subcaption}
\usepackage[utf8]{inputenc} 
\usepackage[T1]{fontenc}    
\usepackage{hyperref}       
\usepackage{url}            
\usepackage{booktabs}       
\usepackage{amsfonts}       
\usepackage{nicefrac}       
\usepackage{microtype}      
\usepackage{xcolor} 
\usepackage{amsmath}
\usepackage{amssymb}
\usepackage{multirow}
\usepackage{algorithm}
\usepackage{algorithmic}
\usepackage{pifont}

\title{Active Source Free Domain Adaptation}

\author{%
  Fan Wang \\
  Shandong University\\
  \texttt{fanwangsail@gmail.com} \\
   \And
  Zhongyi Han\\
  Shandong University \\
   \texttt{hanzhongyicn@gmail.com} \\
  \AND
   Zhiyan Zhang\\
   Shandong University \\
   \texttt{2474101359@qq.com} \\
   \And
   Yilong Yin \\
   Shandong University \\
  \texttt{ylyin@sdu.edu.cn} \\
}

\begin{document}

\maketitle

\begin{abstract}
Source free domain adaptation (SFDA) aims to transfer a trained source model to the unlabeled target domain without accessing the source data. However, the SFDA setting faces an effect bottleneck due to the absence of source data and target supervised information, as evidenced by the limited performance gains of newest SFDA methods. In this paper, for the first time, we introduce a more practical scenario called active source free domain adaptation (ASFDA) that permits actively selecting a few target data to be labeled by experts. To achieve that, we first find that those satisfying the properties of neighbor-chaotic, individual-different, and target-like are the best points to select, and we define them as the minimum happy (MH) points. We then propose minimum happy points learning (MHPL) to actively explore and exploit MH points. We design three unique strategies: neighbor ambient uncertainty, neighbor diversity relaxation, and one-shot querying, to explore the MH points. Further, to fully exploit MH points in the learning process, we design a neighbor focal loss that assigns the weighted neighbor purity to the cross-entropy loss of MH points to make the model focus more on them. Extensive experiments verify that MHPL remarkably exceeds the various types of baselines and achieves significant performance gains at a small cost of labeling.
\end{abstract}

\section{Introduction}

Transferring a trained source model instead of the source data to the unlabeled target domain, source-free domain adaptation (SFDA) has drawn much attention recently. Since it prevents the external leakage of source data, SFDA meets privacy persevering \cite{wang2009survey,kreso2021data}, data security~\cite{thapa2021precision,zhang2018big}, and data silos~\cite{stan2021privacy}. Moreover, it has important potential in many applications, \emph{e.g.,} object detection~\cite{DBLP:conf/aaai/LiCXYYPZ21}, object recognition~\cite{liangjian,li2020model}, and semantic segmentation~\cite{DBLP:conf/wacv/KlingnerTRF22,DBLP:conf/cvpr/LiuZW21}. However, the SFDA setting faces an \emph{effect bottleneck} due to the absence of source data and target
supervised information. In literature, the state-of-the-art A$^2$Net \cite{xia2021adaptive} is a very powerful method, but it only improved the mean accuracy of the pioneering work (SHOT \cite{liangjian}) from 71.8\% to 72.8\% on the challenging Office-Home dataset \cite{venkateswara2017deep}.

 In this paper, we introduce a new setting, \emph{active source free domain adaptation (ASFDA)}, to break through the effect bottleneck. Table~\ref{setting_compare} reports the main differences between active learning (AL), domain adaptation (DA), Active DA, SFDA, and ASFDA. It is balanced that ASFDA permits actively selecting a few samples in the target domain to be labeled by experts without accessing the source data and with distribution shifts across domains. ASFDA simultaneously obeys data security maximization, domain discrepancy impact minimization, labeling cost minimization, and performance benefit maximization. It is also more practical than SFDA because seeking domain experts to annotate a few samples is very common in real-world applications, such as COVID-19 screening~\cite{DBLP:journals/tmi/HanWHLCZWZ20,DBLP:conf/ipmi/HanHLWWY21}, autonomous driving \cite{singh2020deep,dhananjaya2021weather}, and speech recognition \cite{1453593,yuan2019gradient}. 
 \begin{table*}[h]
\centering
	\caption{Comparison of ASFDA and other existing settings (‘$\checkmark$' indicates that the condition is met, ‘$\times$' indicates the opposite, and ‘-' indicates that the condition does not exist).}
	\label{tab:performance_comparison}
    \scalebox{.9}{
    \begin{tabular}{l|c|c|c|c|c}
    \toprule
	   Conditions&AL&DA&Active DA&SFDA&ASFDA\cr
		\midrule
		source data  &- & $\checkmark$&$\checkmark$&$\times$  & $\times$\cr
		\midrule
		a few actively labeled target data (training data) &$\checkmark$ & $\times$&$\checkmark$&$\times$  & $\checkmark$\cr
		\midrule
		domain shift &$\times$ & $\checkmark$&$\checkmark$& $\checkmark$ & $\checkmark$\cr
	\toprule
	\end{tabular}}
	\label{setting_compare}
\end{table*}

Exploring and exploiting the most informative querying samples play key roles in ASFDA, while previous AL and Active DA strategies cannot reach this. In this paper, we find the best informative samples are minimum happy (MH) points that satisfy the properties of neighbor-chaotic, individual-different, and target-like. 1) The property of neighbor-chaotic measures the sample uncertainty through their label-chaotic neighbors identified in high dimensional inherent feature space.
Under domain shift, existing AL \cite{he2019towards,DBLP:conf/ijcnn/WangS14,DBLP:conf/cvpr/JoshiPP09}  and Active DA \cite{DBLP:conf/wacv/SuTSLMC20,DBLP:conf/cvpr/FuC0L21,deheeger2021discrepancy} strategies  are difficult to measure the sample uncertainty due to model miscalibration \cite{DBLP:conf/nips/SnoekOFLNSDRN19} and unavailable source data in ASFDA. 2) The property of individual-different guarantees the sample diversity and further reduces the redundant querying.  Previous methods  \cite{DBLP:conf/iccv/PrabhuCSH21,DBLP:conf/cvpr/SinghDTBDBCVN21} ensure diversity by selecting the samples closest to the centroids or searching a core-set \cite{DBLP:conf/iclr/SenerS18}. However, the selected samples would be easily-adaptive target samples that lie in the source distribution and are unrepresentative of the target domain; 3) The informative samples should be target-like because the representative target samples can improve the generalization of the target domain. Existing Active DA methods \cite{DBLP:conf/wacv/SuTSLMC20,DBLP:conf/cvpr/FuC0L21} select target-like samples by a domain discriminator that requires the source data but which is unavailable in ASFDA. Concerning the exploitation of actively selected samples, most methods of AL \cite{he2019towards,DBLP:conf/iclr/SenerS18,ranganathan2017deep,ash2019deep,DBLP:conf/iclr/AshZK0A20} and Active DA \cite{DBLP:conf/wacv/SuTSLMC20,DBLP:conf/cvpr/FuC0L21,DBLP:conf/iccv/PrabhuCSH21,xie2021active} deem these samples as ordinary labeled samples and use standard supervised losses, but which are difficult to fully exploit the selected samples, resulting in limited performance gains.

\begin{figure}
\centering
\includegraphics[height=5cm,width=13.5cm]{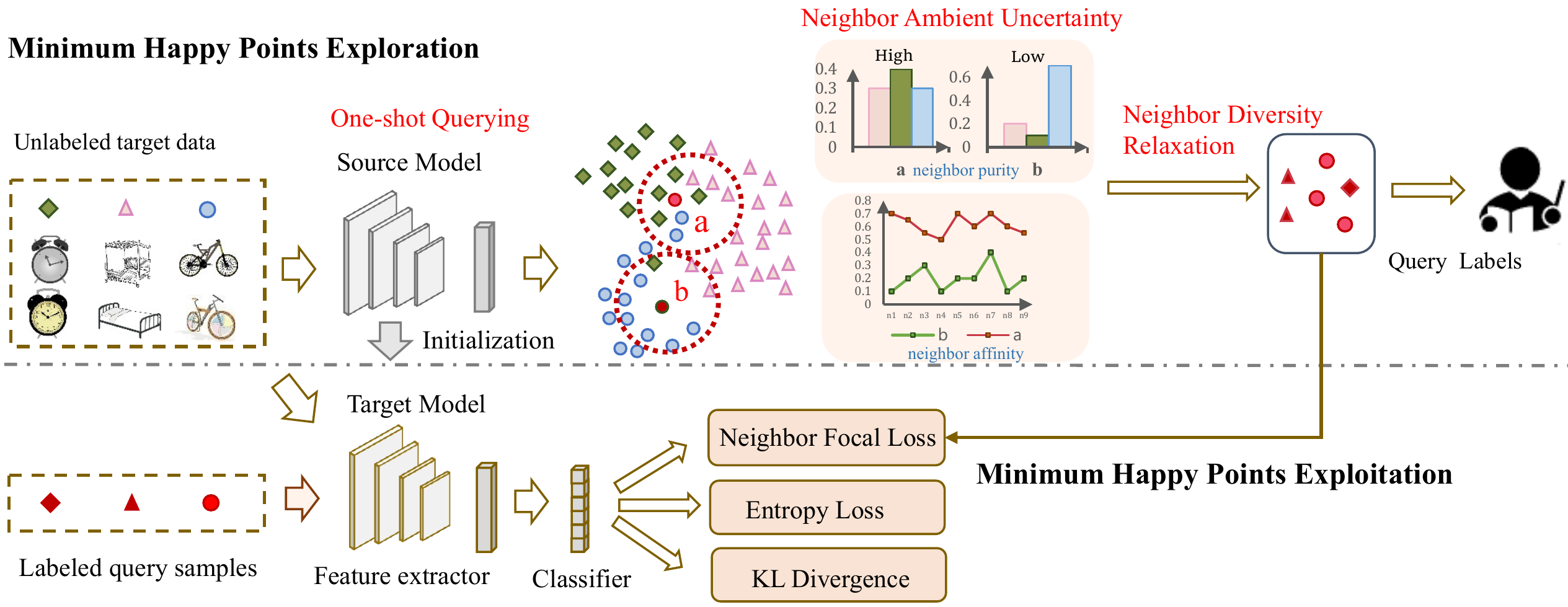}
\caption{The framework of Minimum Happy Points Learning (MHPL).
}
\label{framework}
\end{figure}

We propose the Minimum Happy Points Learning (MHPL) to explore and exploit the informative MH points. The comprehensive framework of MHPL is shown in Fig. \ref{framework}. We propose a novel neighbor ambient uncertainty based on neighbor purity and neighbor affinity to measure the neighbor-chaotic sample uncertainty. Then we propose the neighbor diversity relaxation to guarantee the individual difference. Further, we conduct the one-shot querying, selecting target samples at once based on the source model, as the source model without fine-tuning can better describe the distribution discrepancy across domains and the target-like samples are more likely to be explored. Moreover, the selected samples are fully exploited by a new-designed neighbor focal loss, which assigns the weighted neighbor purity to the cross-entropy loss w.r.t. MH points to make the model focus more on them. 


Our contributions can be summarized as follows: (i) To the best of our knowledge, it is the first time that the new setting of active source free domain adaptation is introduced, which breaks through the effect bottleneck of SFDA; (ii) We find the minimum happy points which enhance the generalization performance on the target domain and achieve successful ASFDA; (iii) We propose a novel MHPL framework to explore and exploit the minimum happy points with the neighbor ambient uncertainty, neighbor diversity relaxation, one-shot querying, and neigh focal loss; (iv) We lay a promising baseline for future works of active source free domain adaptation.

\section{Minimum Happy Points Learning}
\noindent\textbf{Notation.}
In active source free domain adaptation (ASFDA), we can access a source model $h_s : {\mathcal{X}_s} \rightarrow {\mathcal{Y}_s}$ well trained on the source data $D_s$ and an unlabeled target domain $D_t$ = $\{x^i\}_{i=1}^{n_t}$ from different distributions. In the multi-classification task, ${\mathcal{Y}}$ $\in$ $\{1,...,K\}$, and $K$ represents the number of classes. $m$ target samples of the target domain are selected  for querying, $m$ $\ll$ $n_t$. Further, we denote the small selected labeled target data by $D_t^L$ and the unlabeled target data by $D_t^U$. Our method is based on a two-part network: a feature extractor $f$, and a classifier $g$. The feature learned by the feature extractor is denoted as $z_i$ = $f(x_i)$. The goal of ASFDA is to train a target model $h_t$ = $g_t (f_t(x))$ with satisfactory performance while querying as few as labeled data $D_t^L$ from $D_t$ as possible.


\subsection{Minimum Happy Points}
We find the most informative samples for ASFDA are minimum happy points (MH points) that have the characteristics of neighbor-chaotic, individual-different, and target-like at the same time.
\begin{figure}[h]
\begin{minipage}[b]{.33\linewidth}
\centering
\includegraphics[height=3cm,width=3.5cm]{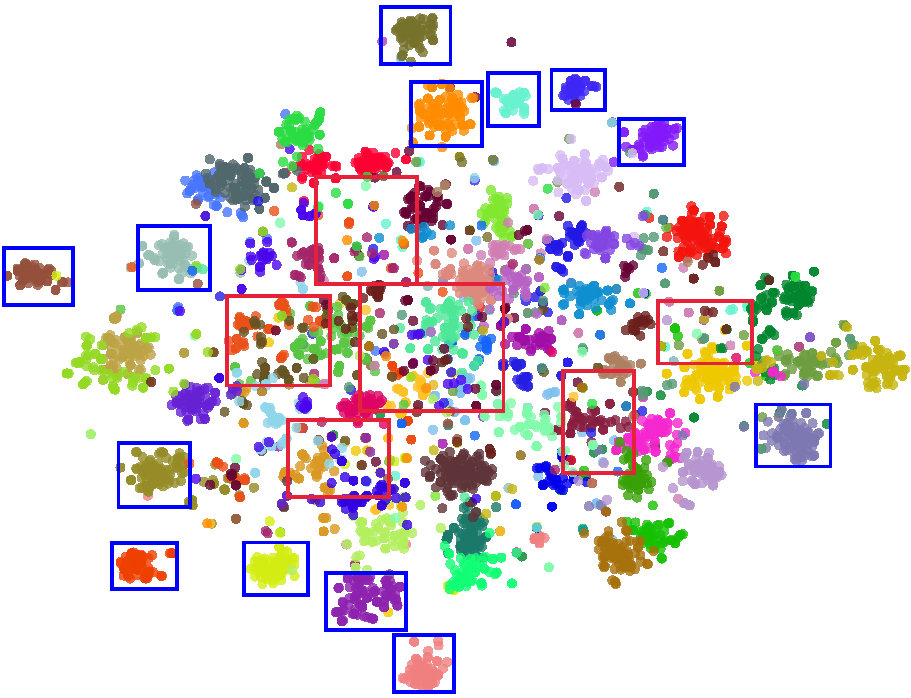}
\subcaption{initial pseudo-labels}
\end{minipage}
\begin{minipage}[b]{.33\linewidth}
\centering
\includegraphics[height=3cm,width=3.5cm]{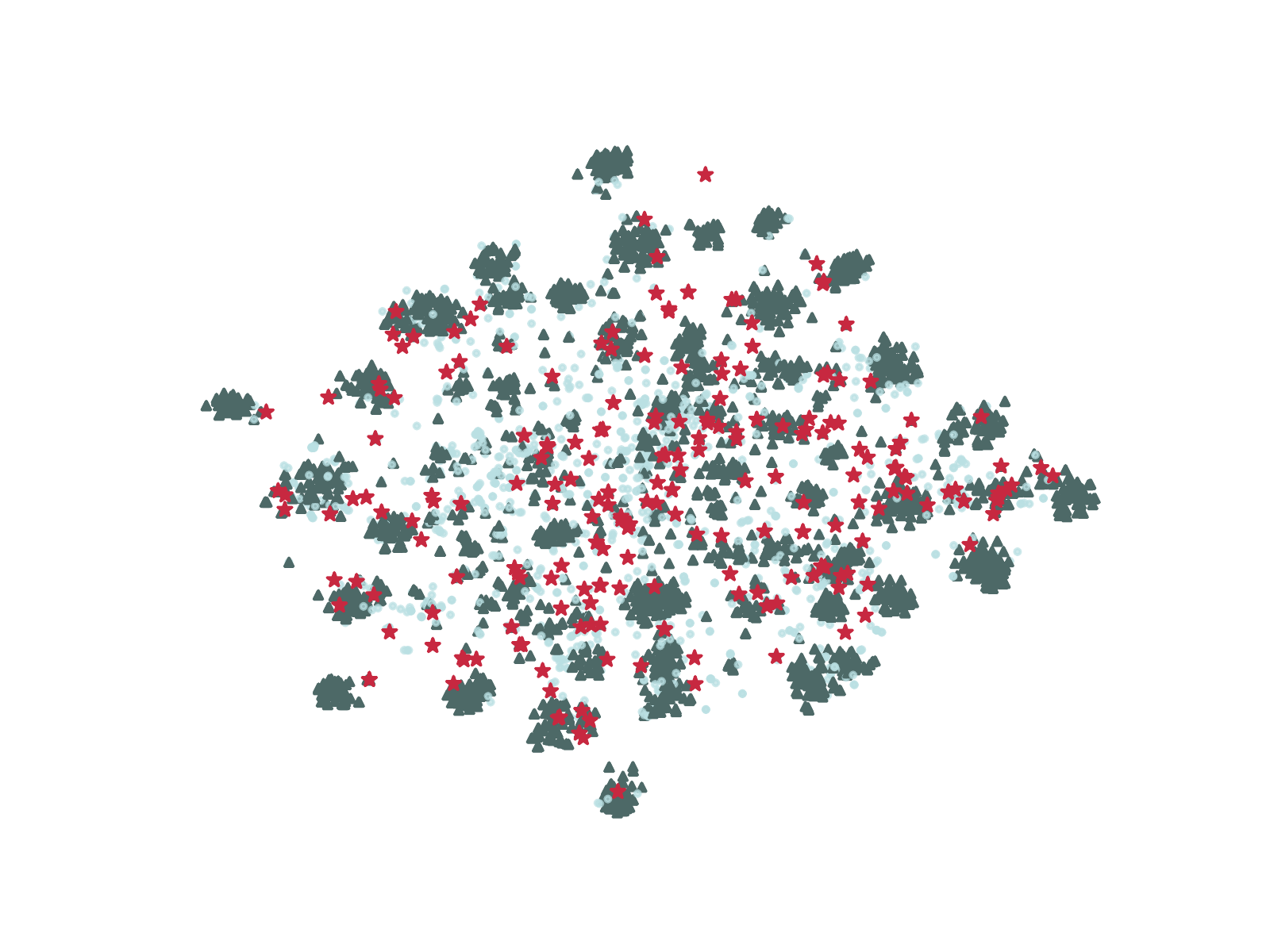}
\subcaption{BVSB \cite{DBLP:conf/cvpr/JoshiPP09}}
\end{minipage}
\begin{minipage}[b]{.33\linewidth}
\centering
\includegraphics[height=3cm,width=3.5cm]{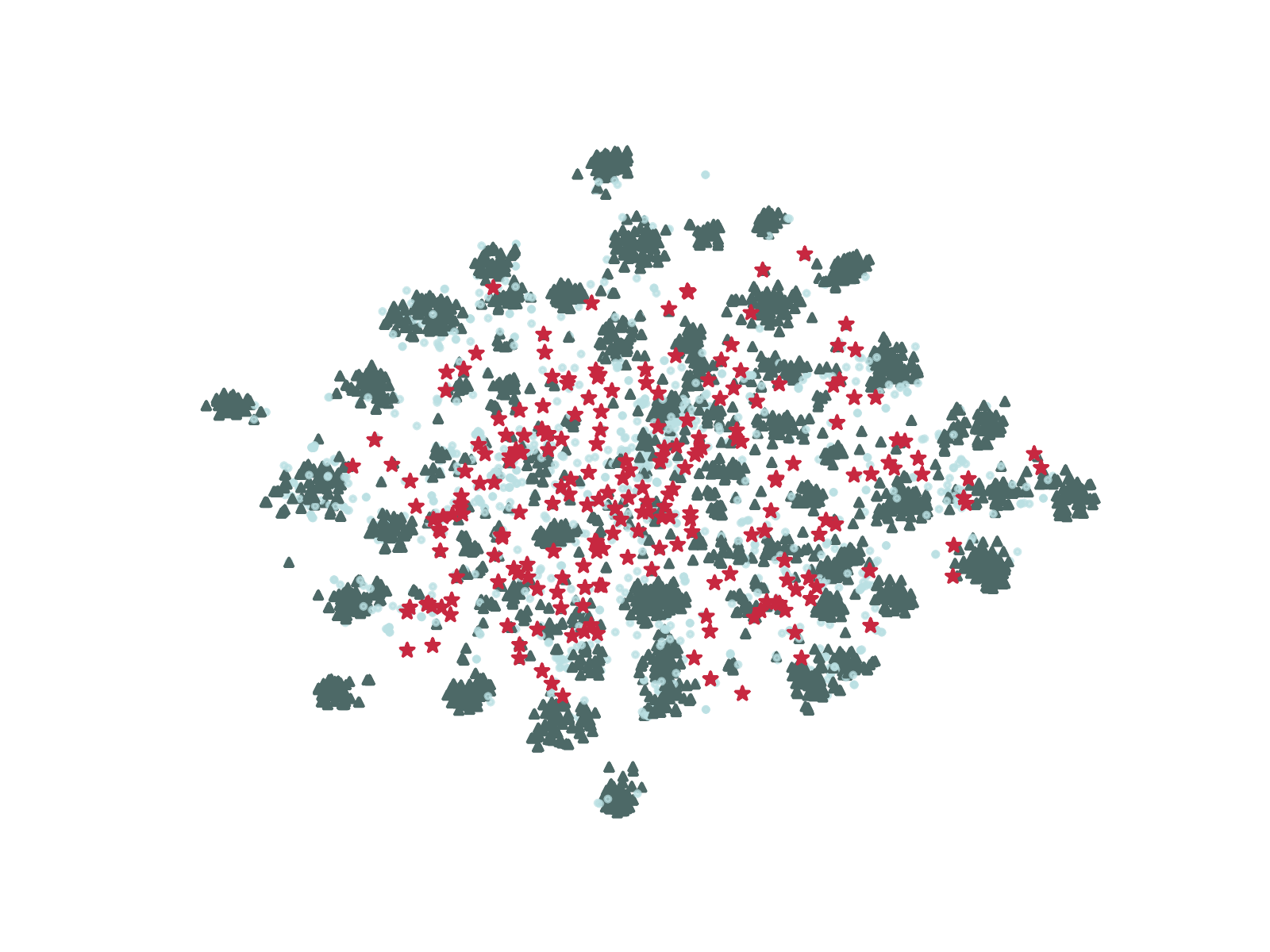}
\subcaption{MHPL}
\end{minipage}
\caption{
Feature visualization for the source model with 5\% actively labeled target data on the Cl$\rightarrow$Pr task. Different colors in (a) represent different classes of pseudo-labels by clustering, where blue blocks include easily-adaptive source-like samples with label-clean neighbors, and red blocks include the hardly-adaptive target-like samples with label-chaotic neighbors.  In (b) and (c), the dark green indicates that the pseudo-label is consistent with the true label, and light blue indicates the opposite. The red stars indicate the selected samples based on BVSB and our MHPL, respectively. 
}
\label{Fig_result_feature_visualization}
\end{figure}

(1) \textbf{Neighbor-chaotic} samples among label-chaotic neighbors are uncertain, meanwhile, their neighbors are heavily uncertain. Annotation of these samples can not only achieve self-correction but also guide the learning of their confusing neighbors, bringing in great performance gains. Further, the sample with chaotic neighbors lies potentially in the boundary of multiple categories, so learning the sample would promote accurate classification boundaries. As shown in Fig. \ref{Fig_result_feature_visualization}, MHPL could explore more samples (red blocks) with label-chaotic neighbors than BVSB \cite{DBLP:conf/cvpr/JoshiPP09} because the strategies based solely on  model uncertainty can only measure the self-uncertainty of samples.

 (2) \textbf{Individual-different} samples are dissimilar to each other, representing sample diversity.  By ensuring the individual difference, we  not only avoid similar instances with high redundancy but also obtain the samples that are representative of the entire target distribution and ensure class balance. It is clear to see in Fig. \ref{Fig_result_feature_visualization}  that both BVSB and MHPL can ensure sample diversity, and MHPL can also avoid selecting outliers, but existing strategies cannot.

 (3) \textbf{Target-like} samples are biased to the target distribution. Following the covariate shift assumption~\cite{david2010impossibility}, the target data $D_t$ can be divided into easily-adaptive source-like instances $D_t^s$ and hardly-adaptive target-like instances $D_t^t$. Selecting samples from $D_t^s$ would lead to limited generalization gain on the target domain since $D_t^s$ represents the source distribution. It is clear to see in Fig.~\ref{Fig_result_feature_visualization} that the samples selected by BVSB have more chances to fall in the blue boxes, which represent the source-like data, while the samples selected by our MHPL are mostly target-like.
 
\subsection{Minimum Happy Points Exploration} 
We propose the neighbor ambient uncertainty, neighbor diversity relaxation, and one-shot querying to explore the neighbor-chaotic, individual-different, and target-like samples.

\noindent\textbf{Neighbor Ambient Uncertainty.} Instead of relying heavily on the self-uncertainty of target samples, the neighbor ambient uncertainty evaluates target sample uncertainty by measuring the neighbor environments they are in. Given a target sample $x$, its neighbor ambient uncertainty $\text{NAU}(x)$ is defined by multiplying the neighbor purity $\text{NP}(x)$ and the neighbor affinity $\text{NA}(x)$:
\begin{equation}  
\begin{aligned}  
&\text{NAU}(x) = \text{NP}(x) * 
\text{NA}(x) \,.\\
\end{aligned} 
\label{Eq6:standard}
\end{equation}
Note that $\text{NP}(x) \ge 0$ and $ 0 \leq \text{NA}(x) \leq 1$. A sample with a high value of $\text{NAU}$ would have noisy and close neighbors, satisfying the neighbor-chaotic characteristic of MH points. As shown in Fig.~\ref{framework}, the sample $a$ has more chances to be selected as an MH point. Next, we first define the concept of neighbor and then introduce the neighbor purity and the neighbor affinity, orderly.

Based on the feature space $\mathcal{F}=\{z_1, z_2,.., z_{n_t}\}$ of target data and the cosine similarity measurement \cite{liangjian}, we define the samples that are close to $x$ in the feature space as the neighbors of $x$:
\begin{equation}  
\begin{aligned}  
&S_N (x) = \{N_1, N_2,..., N_q\} , \text{and} \ S_N^Y (x) = \{Y_{N_1}, Y_{N_2},..., Y_{N_q}\}\,, 
\end{aligned} 
\label{Eq1:neighbor numbers}
\end{equation}
 where $q$ denotes the number of neighbors, $S_N$ represents the neighbor sample space of $x$, and $S_N^Y$ represents the neighbor pseudo-label space of $x$. The pseudo-labels of target samples are obtained by the standard deep clustering~\cite{liangjian}. $Y_{N_q}$ represents the pseudo-label of the neighbor sample $N_q$.

Intuitively, if a sample has the same pseudo-label as its most neighbors, its uncertainty is low. In contrast, the sample in label-chaotic neighbor environments has large uncertainty. Based on this intuitive insight, we design the neighbor purity as follows.


\textbf{Neighbor purity} describes the chaotic degree of neighbor labels around a sample.
For calculating the neighbor purity $\text{NP}(x)$, we first establish the neighbor class probability distribution space $S_N^p(x)$:
\begin{equation}  
\begin{aligned}  
&S_N^p(x) = \{p_1, p_2, p_3,..., p_k\}\,,\\
\end{aligned} 
\label{Eq2:neighbor probability}
\end{equation}
where $p_k$ represents the proportion of samples labeled as class $k$ in $q$ neighbors. Take samples $a$ and $b$ in the Fig. \ref{framework} as examples, $S_N^p (a) = [0.3,0.4,0.3]$, $S_N^p (b) = [0.2,0.1,0.7]$.

Then the neighbor purity of sample $x$ is measured by the neighbor entropy:
\begin{equation}  
\begin{aligned}  
&{\text{NP}}(x) = - {\sum}_{k=1}^K p_k \text{log} p_k, \ \text{\text{s.t.}} \ p_k \in S_N^p (x) \,,
\end{aligned} 
\label{Eq3:neighbor purity}
\end{equation}
The sample with a high value of $\text{NP}$ has label-noisy neighbors, and such a sample is easy to be selected as an uncertain one. 
However, only using neighbor purity would select outliers, resulting in biased estimation, so we propose the neighbor affinity as a strong complement to neighbor purity.

\textbf{Neighbor affinity} describes how close a sample is to its neighbors. To measure the close degree, we first define the neighbor similarity space $S_N^s$ for each sample by
\begin{equation}  
\begin{aligned}  
&S_N^s (x)= \{S_{N_1}, S_{N_2},..., S_{N_p}\}\,,
\end{aligned} 
\label{Eq4:neighbor distance}
\end{equation}
where $S_{N_1}$ represents the cosine similarity between $x$ and its neighbor $N_1$. Further, the neighbor affinity is measured by the average similarity between $x$ and its neighbors:
\begin{equation}  
\begin{aligned}  
&\text{NA}(x) = \frac{S_{N_1} + S_{N_2} + ... + S_{N_q}}{q},  \ \text{s.t.} \ S_{N_q} \in S_N^s (x) \,.
\end{aligned} 
\label{Eq5:neighbor avg distance}
\end{equation}
The more low is the neighbor affinity, the more far is the sample to its neighbors, and such that it is more likely to be outliers because outliers do not have compact neighbors.

The advantages of NAU are two-fold. Firstly, previous SFDA works~\cite{yang2021exploiting,yang2021generalized} have proved that the target samples' deep features extracted by the source model can still form clean clusters under domain shift to a certain extent. Fig.~\ref{Fig_result_feature_visualization} shows the visualization of target features and further verifies this phenomenon. NAU measures the sample uncertainty from its neighbors that would be better than the strategies that rely solely on the miscalibrated uncertainty of model predictions~\cite{he2019towards,DBLP:conf/ijcnn/WangS14,DBLP:conf/cvpr/JoshiPP09}. Secondly, NAU has the advantages of ensemble learning \cite{zhou2012ensemble} and lets numerous neighbor crowds calibrate the individual uncertainty and improve the fault tolerance of target sample selection.

\begin{algorithm}[tb]
\caption{Neighbor Diversity Relaxation.}
\label{alg:algorithm}
\textbf{Require}: target data $D_t$, labeled target set $D_t^L=\emptyset$, $m$;
\begin{algorithmic}[1] 
\STATE Sort the samples in $D_t$ in reverse order by the value of $\text{NAU}$, and Let $i$ = 0;
\WHILE{$\text{Length}(D_t^L)$ $\leq$ $m$}
\STATE Select the candidate sample $x_i$ and obtain its nearest neighbor $N_{x_i}^o$,
\STATE \textbf{if} {$N_{x_i}^o$ $\notin$ $D_t^L$ } \textbf{then} $D_t^L$ $ \leftarrow $ $x_i$,
\STATE \textbf{else}  Skip the selection of sample $x_i$,
\ENDWHILE
\end{algorithmic}
\label{NDR_al}
\end{algorithm}

\noindent\textbf{Neighbor Diversity Relaxation.} 
We propose the neighbor diversity relaxation (NDR) to guarantee individual differences in candidate samples. In existing methods \cite{DBLP:conf/iccv/PrabhuCSH21,DBLP:conf/cvpr/SinghDTBDBCVN21}, sample diversity is ensured by choosing samples closest to cluster centers, which ignores domain shift and selects uninformative and source-like samples. Instead, NDR guarantees sample diversity by performing nearest neighbor relaxation on candidate samples with high neighbor ambient uncertainty. The main procedure of NDR is shown in algorithm~\ref{NDR_al}. For a candidate sample $x$ with a high $\text{NAU}(x)$, we first evaluate its nearest neighbor sample $N_{x_i}^o$ before putting it in $D_t^L$. If $N_{x_i}^o$ has been put in the $D_t^L$, then $x_i$ has no chance to be selected for expert annotation. In this way, those redundant samples with similar features would have few chances to be selected, ensuring the individual difference and sample diversity.
 
 \noindent\textbf{One-shot Querying.} In order to explore target-like samples, we perform one-shot querying, selecting samples at once according to the raw source model. The reasons are two-fold. For one thing, without the source data, we can measure the domain discrepancy by observing how the target data behaves on the  untrained source model. The samples that can be easily classified by the source model are commonly source-like instances, which mainly occur in label-clean clusters \cite{yang2021exploiting,Tang_2020_CVPR}. Otherwise, the samples with label-chaotic neighbors are severely misclassified and target-like instances which can be effectively explored by our proposed criterion. For another, as the training model is biased toward the target domain gradually, it is difficult to determine if the samples are source-like or target-like.

\subsection{Minimum Happy Points Exploitation}

\noindent\textbf{Neighbor Focal Loss} is designed to make the model focus more on chaotic and representative MH points to ensure the generalization of the target domain. NF loss is inspired by, but not the same as, focal loss \cite{lin2017focal}. For one side, focal loss aims to alleviate the overfitting problem of cross-entropy loss for imbalanced object detection datasets, while NF loss is proposed to fully exploit the MH points. For another, the focal loss assigns the weights from the miscalibrated model's unreliable predictions to the hard samples, while NF loss assigns the neighbor purity and a larger weight $\alpha$ to the MH points:
\begin{equation}  
\begin{aligned}  
&L_{NF} (f_{t};{\mathcal{D}_{t}^L})= - {\mathbb{E}}_{x \in {\mathcal{D}}_{t}^L}{\sum}_{k=1}^K  \alpha\text{NP}(x) l_k \text{log}({\delta}_k (f_{t}(x)))\,,\\
\end{aligned} 
\label{Eq8:NCEloss_L_t}
\end{equation}
where $\delta$ is the softmax function,
and $l_k$ is ‘1’ for the ground-truth class. Focusing on informative MH points can increase the importance of active labeled samples and further avoid the model to overfit on the samples with wrong pseudo-labels. And assigning neighbor purity could penalty the mistakes on neighbor-chaotic samples that are more important than ordinary samples for generalization. 


Meanwhile, the NF loss assigns a smaller weight $\beta$ to the rest target samples $D_t^U$. NF loss uses their pseudo-labels from clustering and makes the model do not tilt toward them during training:
\begin{equation}  
\begin{aligned}  
&L_{NF}(f_{t};{\mathcal{D}_{t}^U}) = - {\mathbb{E}}_{x \in {\mathcal{D}}_{t}^U}{\sum}_{k=1}^K \beta l_k \text{log}({\delta}_k (f_{t}(x)))  \,.\\
\end{aligned} 
\label{Eq8:NCEloss_L_u}
\end{equation}

In summary, NF loss pays the more attention to MH points and the low attention to non-MH points:
\begin{equation}  
\begin{aligned}  
&L_{NF} (f_{t};{\mathcal{D}_{t}})= L_{NF}(f_{t};{\mathcal{D}_{t}^L}) + L_{NF}(f_{t};{\mathcal{D}_{t}^U})   \,.\\
\end{aligned} 
\label{Eq9:NCEloss}
\end{equation}
\noindent\textbf{Entropy loss}  and \noindent\textbf{KL divergence} are introduced to guarantee the unambiguous and balanced classes \cite{bridle1991unsupervised,hu2017learning}, which has been widely used in clustering \cite{ghasedi2017deep,krause2010discriminative}, and several DA works \cite{liangjian,Tang_2020_CVPR,shi2012information,huang2021model}:
\begin{equation}  
\begin{aligned}  
&L_{ent}(f_{t};{\mathcal{D}_{t}})=-{\mathbb{E}}_{x \in {\mathcal{D}}_{t}}{\sum}_{k=1}^K {\delta}_k(f_{t}(x))\text{log}({\delta}_k (f_{t}(x)))\,,\\
&L_{div}(f_{t};{\mathcal{D}_{t}})=-{\mathbb{E}}_{x \in {\mathcal{D}}_{t}}{\sum}_{k=1}^K \text{KL}(\hat p_k||q_k), \text{and}  \ q_{\{k=1,...,K\}} = \frac{1}{K} \,,\\
\end{aligned} 
\label{Eq10:entropy_loss}
\end{equation}
where  $\hat p_k =\frac{1}{n_t} \sum\delta(f_{t}(x))^{(k)}$ is the mean prediction of the $k$-th target data.

\begin{algorithm}[tb]
\caption{Minimum Happy Points Learning.}
\label{alg:algorithm}
\textbf{Require}: source model $h_s $, target data $D_t$, labeled target set $D_t^L=\emptyset$, maximum number of epochs $E$, trade-off parameters $\alpha$, $\beta$, $m$;
\begin{algorithmic}[1] 
\STATE Initialize the target model $h_t$ with $h_s$ and obtain pseudo-labels of target data based clustering;
\STATE $\forall x \in D_t$, compute  $\text{NP}(x)$ and  $\text{NA}(x)$ to serve as $\text{NAU}(x)$ with Eq. (\ref{Eq6:standard} - \ref{Eq5:neighbor avg distance});
\STATE $D_t^L$ $\leftarrow$ select $m$ samples with $\text{NAU}$ and algorithm \ref{NDR_al};
\STATE Let epoch $= 1$, $iter\_num =0$;
\WHILE{epoch $\leq$ $E$}
\STATE Obtain pseudo-labels of $D_t^U$ from deep clustering;
\WHILE{$iter\_num < n_b$}
\STATE{Calculate} neighbor focal loss, entropy loss  and KL divergence with Eq. (\ref{Eq8:NCEloss_L_t}- \ref{Eq10:entropy_loss}); \\
\STATE Update model $f_t$ via Eq. (\ref{Eq12:Totalloss}).
\ENDWHILE
\ENDWHILE

\end{algorithmic}
\label{ALG1}
\end{algorithm}
To conclude, the workflow of MHPL is illustrated in Algorithm \ref{ALG1}. The final objective can be stated as\begin{equation}
\begin{aligned}  
&L = L_{NF} + L_{ent}  + L_{div}\,.
\end{aligned} 
\label{Eq12:Totalloss}
\end{equation}
\section{Experiments}

\noindent\textbf{Benchmarks.} 
We adopt three benchmark datasets. Office-31 \cite{saenko2010adapting} is a small-scale DA dataset with 31 classes and 3 domains: \textbf{A}mazon, \textbf{D}slr, and \textbf{W}ebcom. Office-Home \cite{venkateswara2017deep} is a more challenging DA dataset with 65 classes and 4 domains: \textbf{Ar}tistic images, \textbf{Cl}ip Art images, \textbf{Pr}oduct images, and \textbf{Re}al-world images. VisDA-2017 \cite{peng2017visda} is a large simulation-to-real dataset with 12 classes. 

\noindent\textbf{Baselines.} We construct three types of methods as baselines.
(i) Source free domain adaptation: \textbf{SFDA} \cite{kim2020towards}, \textbf{SHOT} \cite{liangjian},  \textbf{MA} \cite{li2020model},  \textbf{HCL} \cite{huang2021model},  \textbf{CPGA} \cite{qiu2021source}, \textbf{NRC} \cite{yang2021exploiting}, and \textbf{A$^2$Net}  \cite{xia2021adaptive}. (ii)  Active learning:
(1) \textbf{Random}: random samples.
(2) \textbf{Entropy} \cite{DBLP:conf/ijcnn/WangS14}: samples with highest entropy.
(3) \textbf{BVSB} \cite{DBLP:conf/cvpr/JoshiPP09}: samples with the smallest difference between top-2 class probabilities.
(4) \textbf{LC} (least confidence) \cite{he2019towards}: samples with smallest probability.
(5) \textbf{CoreSet} \cite{DBLP:conf/iclr/SenerS18}: samples selected by a set-cover problem.
(6) \textbf{CTC}: samples that are closest to clustering centers.
(7) \textbf{BADGE} \cite{DBLP:conf/iclr/AshZK0A20}: construct diverse batches by running KMeans++ \cite{arthur2006k}.
(iii) Active DA: \textbf{AADA} \cite{DBLP:conf/wacv/SuTSLMC20},  \textbf{TQS} \cite{DBLP:conf/cvpr/FuC0L21}, \textbf{CLUE} \cite{DBLP:conf/iccv/PrabhuCSH21}, and \textbf{EADA} \cite{DBLP:journals/corr/abs-2112-01406}.

\noindent\textbf{Implementation.} We report the main results upon the backbone of ResNet-50 \cite{he2016deep} for Office-Home and Office-31, as well as ResNet-101 for VisDA-2017. We adopt the same network architecture as SHOT \cite{liangjian}. We conduct SGD with momentum 0.9 and batch size of 64 for all datasets. The learning rate is set to 1e-2 for Office-31 and Office-Home, and 1e-3 for VisDA-2017. For the number of neighbors $q$, we set nine for Office-31, 20 for Office-Home, and five for VisDA-2017. Further, we set $\alpha$ = 3, $\beta$ = 0.3 for all datasets. The more implementation details are in Appendix C. The full results of other backbones and the sensitivity analysis of the hyperparameters are in Appendix D.
\subsection{Main Results}
\begin{table*}
\centering
	\caption{Accuracy (\%) on Office-Home (ResNet50) under different settings with 5\% labeled target samples ("SF" in tables denotes source data free, \emph{i.e.,} adaptation without source data).}
	\label{tab:performance_comparison}
    \scalebox{.55}{
    \renewcommand{\arraystretch}{1.2}
    \begin{tabular}{l|c|c|ccccccccccccccc}
    \toprule
    Categories&Method&SF&Ar$\rightarrow$Cl&Ar$\rightarrow$Pr&Ar$\rightarrow$Re&Cl$\rightarrow$Ar&Cl$\rightarrow$Pr&Cl$\rightarrow$Re&Pr$\rightarrow$Ar&Pr$\rightarrow$Cl&Pr$\rightarrow$Re&Re$\rightarrow$Ar&Re$\rightarrow$Cl&Re$\rightarrow$Pr&Avg\cr
    \midrule
    None& Source-only&$\checkmark$&45.5&68.4&75.2&53.4&63.7&65.6&52.4&41.0&73.6&65.9&46.3&78.2&60.8\cr
    \midrule
    \multirow{4}{*}{SFDA} 
       & SFDA \cite{kim2020towards}  & $\checkmark$ & 48.5 & 71.3 & 75.6 & 63.9 & 69.0 & 72.1 & 62.4 & 43.5 & 76.0 & 70.4 & 50.1 & 76.1 & 64.9\\ 
        & CPGA \cite{qiu2021source} & $\checkmark$ & 59.3 & 78.1 & 79.8 & 65.4 & 75.5 & 76.4 & 65.7 & 58.0 & 81.0 & 72.0 & 64.4 & 83.3 & 71.6\\
       & SHOT \cite{liangjian} & $\checkmark$ & 57.1 & 78.1 & 81.5 & 68.0 & 78.2 & 78.1 & 67.4 & 54.9 & 82.2 & 73.3 & 58.8 & 84.3 & 71.8\\ 
       & NRC \cite{yang2021exploiting} &$\checkmark$ & 57.7 & 80.3 & 82.0 & 68.1 & 79.8 & 78.6 & 65.3 & 56.4 & 83.0 & 71.0 & 58.6 & 85.6 & 72.2\\
       & A$^2$Net \cite{xia2021adaptive} &$\checkmark$ & 58.4 & 79.0 & 82.4 & 67.5 & 79.3 & 78.9 & 68.0 & 56.2 & 82.9 & 74.1 & 60.5 & 85.0 & 72.8\\
    \midrule
    \multirow{4}{*}{Active DA}
	    &AADA \cite{DBLP:conf/wacv/SuTSLMC20}  &$\times$  & 56.6 & 78.1 & 79.0 & 58.5 &72.7& 71.0 &60.1 &53.1 &77.0 &70.6 &57.0& 84.5 & 68.3\cr
		 &TQS \cite{DBLP:conf/cvpr/FuC0L21}  & $\times$& 58.6 & 81.1 & 81.5 & 61.1 & 76.1 &73.3 & 61.2 & 54.7 & 79.7 & 73.4 & 58.9 &86.3 & 72.5\cr
		  &Clue \cite{DBLP:conf/iccv/PrabhuCSH21}  &$\times$ &58.0 & 79.3 & 80.9 & 68.8 & 77.5 & 76.7 &66.3 & 57.9 & 81.4 & 75.6 & 60.8 & 86.3 & 72.5\cr
		  &EADA \cite{DBLP:journals/corr/abs-2112-01406}  & $\times$& 63.6& 84.4 & 83.5 &70.7 & 83.7 & 80.5 & 73.0 &63.5 & 85.2 & 79.4 & 65.4 & 88.6 & 76.7\cr
     \midrule
     \multirow{8}{*}{ASFDA} 
	   	&Base&$\checkmark$&57.2&78.5&81.5&68.5&79.1&78.6&67.5&56.3&82.2&73.7&58.5&83.6&72.1\cr
	    &Random  &$\checkmark$&63.8&81.4&83.9&71.3&82.2&81.4&68.8&62.4&83.3&76.1&63.8&85.8&75.2\cr
		&CTC &$\checkmark$  & 60.8 & 78.7 & 82.2 & 69.3 & 79.2 & 79.8 & 68.6 & 59.4& 82.2& 74.6 & 61.7 & 84.4 & 73.4 \cr
		&CoreSet \cite{DBLP:conf/iclr/SenerS18} &$\checkmark$  &61.8  &81.8& 83.3 &71.1&82.9&81.6&70.7&60.5 &84.7&76.1&61.7&86.1& 75.2\cr
		&BADGE \cite{DBLP:conf/iclr/AshZK0A20} &$\checkmark$  &62.4  &82.7& 83.9 &71.5&83.0&81.8&71.2&62.7 &84.6&76.2&62.9&87.8& 75.9\cr
		&Entropy \cite{DBLP:conf/ijcnn/WangS14}&$\checkmark$  &65.0&84.0&85.9&71.8&83.8&82.6&70.7&63.8&85.1&77.8&64.1&88.1&76.9\cr
		&BVSB \cite{DBLP:conf/cvpr/JoshiPP09}&$\checkmark$  &64.8&84.4&85.5&72.0&83.2&83.4&70.4&63.9&85.0&77.5&65.0&88.1&76.9\cr
		&LC \cite{he2019towards}&$\checkmark$  &65.0&84.0&85.4&72.1&83.0&82.8&71.0&64.9&85.1&78.0&64.8&87.9&77.0\cr
		\cmidrule{2-16}
		&\textbf{MHPL}& $\checkmark$ &\textbf{67.8}&\textbf{85.9}&\textbf{86.1}& \textbf{72.8}&\textbf{87.1}&\textbf{84.5}&\textbf{72.5}&\textbf{67.6}&\textbf{86.2}&\textbf{79.7}&\textbf{68.6}&\textbf{89.9}&\textbf{79.1}\cr
		\toprule
	\end{tabular}}
	\label{Result_Office_Home}
\end{table*}
\begin{table*}
\centering
	\caption{Accuracy (\%) on VisDA-2017 (ResNet-101) and Office-31 (ResNet-50) with 5\% labeled target samples ("SF"  in tables denotes source data free, \emph{i.e.,} adaptation without source data).  }
	\label{tab:performance_comparison}
    \scalebox{.65}{
    \begin{tabular}{l|c|c|c|cccccccc}
    \toprule
	   Categories&Method&SF&VisDA-2017&A$\rightarrow$D&A$\rightarrow$W&D$\rightarrow$A&D$\rightarrow$W&W$\rightarrow$A&W$\rightarrow$D&Avg\cr
		\midrule	None&Source-only&$\checkmark$&50.0&79.3&76.7&61.8&96.7&63.0&98.8&79.4\cr
		\midrule
		\multirow{4}{*}{SFDA} 
          & SHOT \cite{liangjian} & $\checkmark$ & 82.4 & 94.0 & 90.1 &74.7& 98.4 & 74.3 &99.9 & 88.6\cr
          & CPGA \cite{qiu2021source}& $\checkmark$ & 86.0 & 94.4 &94.1 & 76.0 &98.4 & 76.6 & 99.8 & 89.9\cr
          & HCL \cite{huang2021model}& $\checkmark$ & 83.5 & 90.8 &91.3 & 72.7 & 98.2 & 72.7 & \textbf{100.0} & 87.6\cr
          & NRC \cite{yang2021exploiting}&$\checkmark$ & 85.9 & 96.0 & 90.8 & 75.3 & 99.0 & 75.0 & \textbf{100.0} & 89.4\cr
          & A$^2$Net \cite{xia2021adaptive}&$\checkmark$ & 84.3 & 94.5 & 94.0 & 76.7 & 99.2 & 76.1 & \textbf{100.0} & 90.1\cr
        \midrule
        \multirow{4}{*}{Active DA}
	    &AADA \cite{DBLP:conf/wacv/SuTSLMC20} &$\times$  & - & 89.2 & 87.3 & 78.2 &99.5 & 78.7 & \textbf{100.0} &88.8\cr
		&TQS \cite{DBLP:conf/cvpr/FuC0L21} & $\times$& - & 92.8 & 92.2 & 80.6 & \textbf{100.0} &80.4 & \textbf{100.0} & 91.1\\
		&Clue \cite{DBLP:conf/iccv/PrabhuCSH21}&$\times$ & - & 92.0 & 87.3 & 79.0 & 99.2 & 79.6 & 99.8 & 89.5\\
		&EADA \cite{DBLP:journals/corr/abs-2112-01406} & $\times$&- & 97.7  & 96.6& 82.1 & \textbf{100.0} & 82.8 &\textbf{100.0} & 93.2 \\
        \midrule
        \multirow{8}{*}{ASFDA} 
	   	&Base&$\checkmark$&83.3&93.6&90.6&75.5&98.9&75.4&99.8&89.0\cr
	    &Random&$\checkmark$  &85.1&94.2&94.8&78.2&98.9&77.4&99.6&90.5\cr
		&CTC&$\checkmark$  &84.0&94.0&90.1&77.1&98.6&76.1&99.8&89.3\cr
		&CoreSet \cite{DBLP:conf/iclr/SenerS18}&$\checkmark$  &85.9&93.6&91.8&78.6&99.1&78.0&99.8&90.2\cr
		&BADGE \cite{DBLP:conf/iclr/AshZK0A20}&$\checkmark$  &86.0&95.2&91.5&78.2&99.1&78.2&99.8&90.3\cr
		&Entropy \cite{DBLP:conf/ijcnn/WangS14}&$\checkmark$  &86.7&96.0&94.0&80.5&99.0&79.3&\textbf{100.0}&91.5\cr
		&BVSB\cite{DBLP:conf/cvpr/JoshiPP09}&$\checkmark$  &86.5&95.4&94.0&80.3&99.1&80.1&\textbf{100.0}&91.5\cr
		&LC \cite{he2019towards}&$\checkmark$  &86.7&96.2&95.2&80.1&99.1&79.7&\textbf{100.0}&91.7\cr
		\cmidrule{2-11}
		&\textbf{MHPL}& $\checkmark$ &\textbf{91.2}&\textbf{98.6}&\textbf{96.7}& \textbf{82.3}&99.3&\textbf{83.9}&\textbf{100.0}&\textbf{93.5}\cr
		\toprule
	\end{tabular}}
	\label{Result_Office31}
\end{table*}
\begin{table*}[!h]
\centering
	\caption{Accuracy (\%) on VisDA-2017 (ResNet-50) with 5\% labeled target samples.}
	\label{tab:performance_comparison}
    \scalebox{.8}{
    \begin{tabular}{l|c|c|c|c|c}
    \toprule
	   Method&AADA \cite{DBLP:conf/wacv/SuTSLMC20}&TQS \cite{DBLP:conf/cvpr/FuC0L21}&Clue\cite{DBLP:conf/iccv/PrabhuCSH21}&EADA \cite{DBLP:journals/corr/abs-2112-01406}&MHPL\cr
		\midrule
		Acc ($\%$) &80.8 $\pm$ 0.4 &83.1 $\pm$ 0.4&85.2 $\pm$ 0.4& 88.3 $\pm$ 0.1 &\textbf{89.5  $\pm$ 0.1}\cr
	\toprule
	\end{tabular}}
	\label{Result_visda_resnet50}
\end{table*}

\noindent\textbf{Results on object recognition.} Our method MHPL significantly outperforms existing SFDA methods, successfully breaking through the effect bottleneck of SFDA with limited annotations. Firstly, MHPL achieves the state-of-the-art on the ASFDA setting. As shown in Table \ref{Result_Office_Home}, MHPL surpasses on average by 6.3\% over the state-of-the-art SFDA method A$^2$Net on Office-Home with only 5\% of labeled target data. Especially in challenging tasks, MHPL achieves significant improvements, \emph{e.g.,} the accuracy of MHPL is 11.4\% and 9.5\% higher than that of A$^2$Net on tasks Ar$\rightarrow$Cl and Pr$\rightarrow$Cl, respectively. As shown in Table \ref{Result_Office31}, the accuracy of MHPL also remarkably outperforms all SFDA baselines on VisDA-2017 and Office-31. Secondly, MHPL can better explore and exploit MH points and maximize the performance gains compared to existing active learning baselines. As shown in Tables~\ref{Result_Office_Home}, \ref{Result_Office31}, MHPL outperforms all existing active learning strategies on all tasks, especially for challenging tasks. Finally, as shown in Tables~\ref{Result_Office_Home}, \ref{Result_Office31}, \ref{Result_visda_resnet50}, it is very interesting that MHPL without accessing the source data can outperform all active domain adaptation methods, showing the practical significance of ASFDA. 

\subsection{Analysis}
\begin{figure}
\begin{minipage}[b]{.33\linewidth}
\centering
\includegraphics[height=2.8cm,width=3.1cm]{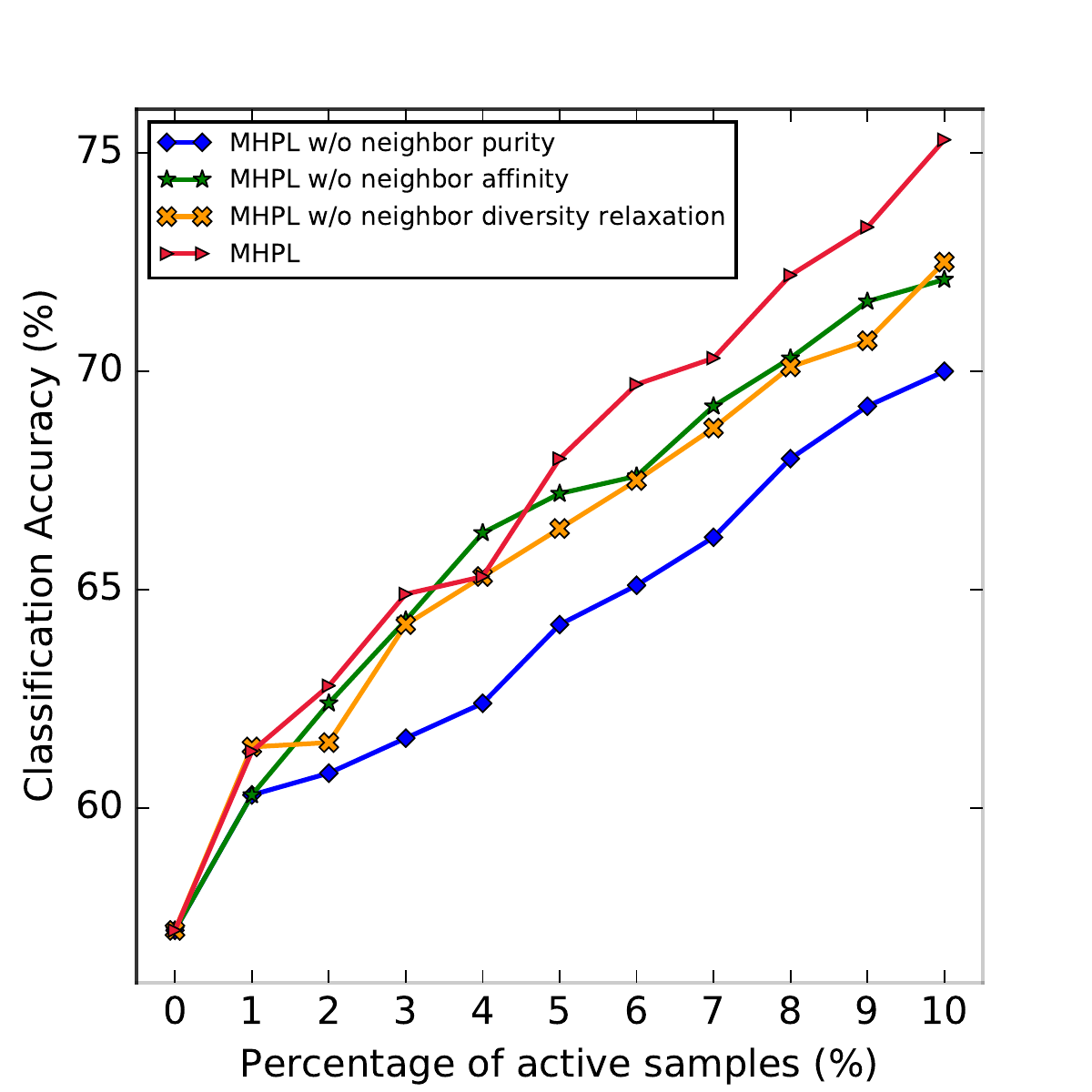}
\subcaption{Ar$\rightarrow$Cl}
\end{minipage}
\begin{minipage}[b]{.33\linewidth}
\centering
\includegraphics[height=3.0cm,width=3.5cm]{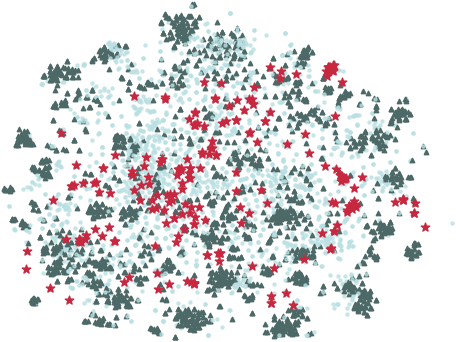}
\subcaption{Ar$\rightarrow$Cl without NDR}
\label{FCS_b}
\end{minipage}
\begin{minipage}[b]{.33\linewidth}
\centering
\includegraphics[height=3.0cm,width=3.5cm]{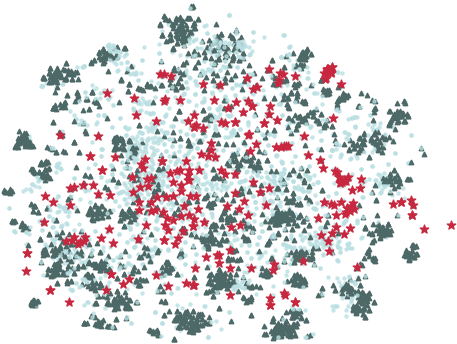}
\subcaption{Ar$\rightarrow$Cl}
\label{FCS_c}
\end{minipage}

\begin{minipage}[b]{.25\linewidth}
\centering
\includegraphics[height=2.8cm,width=3.1cm]{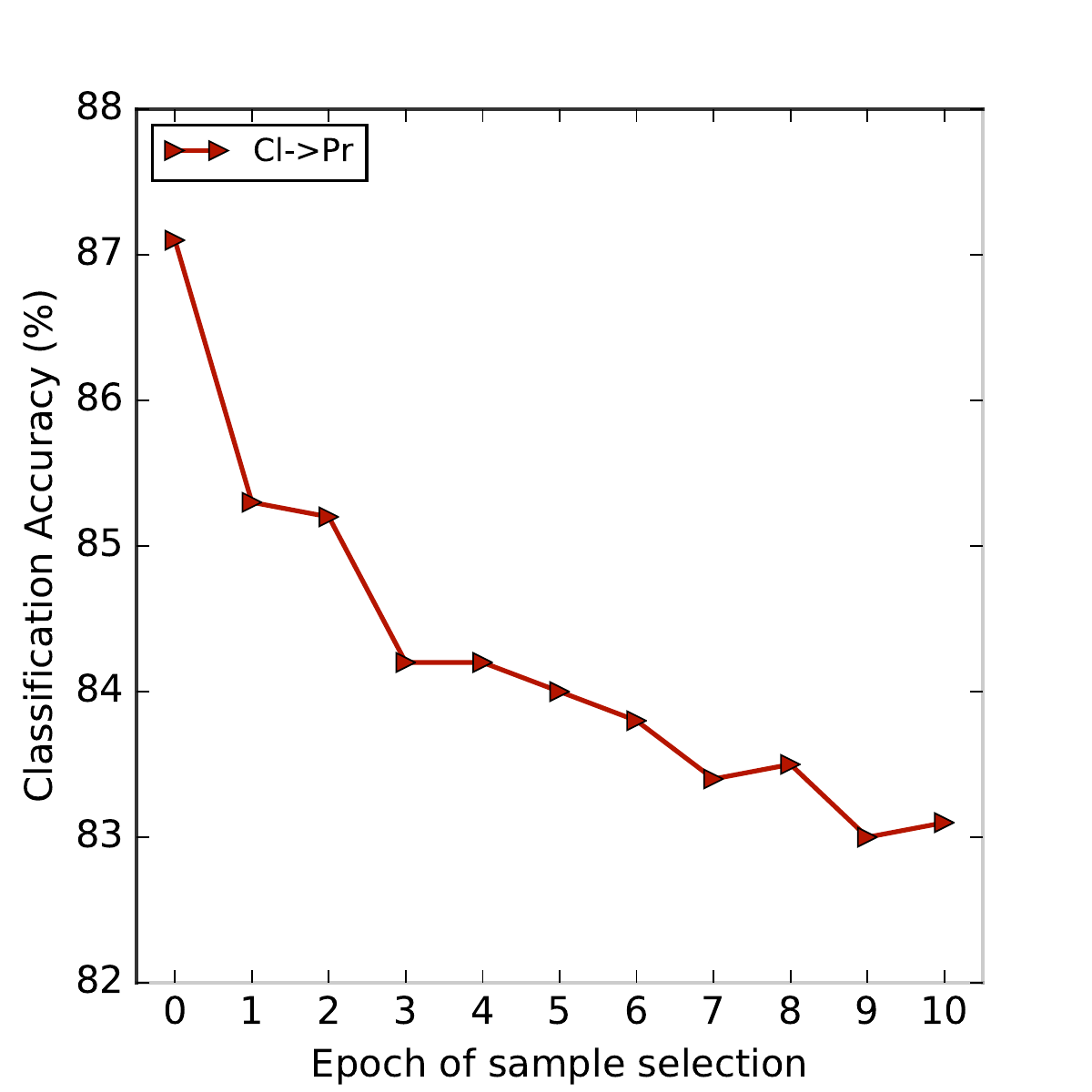}
\subcaption{Cl$\rightarrow$Pr}
\label{FCS_a}
\end{minipage}%
\begin{minipage}[b]{.25\linewidth}
\centering
\includegraphics[height=2.8cm,width=3.1cm]{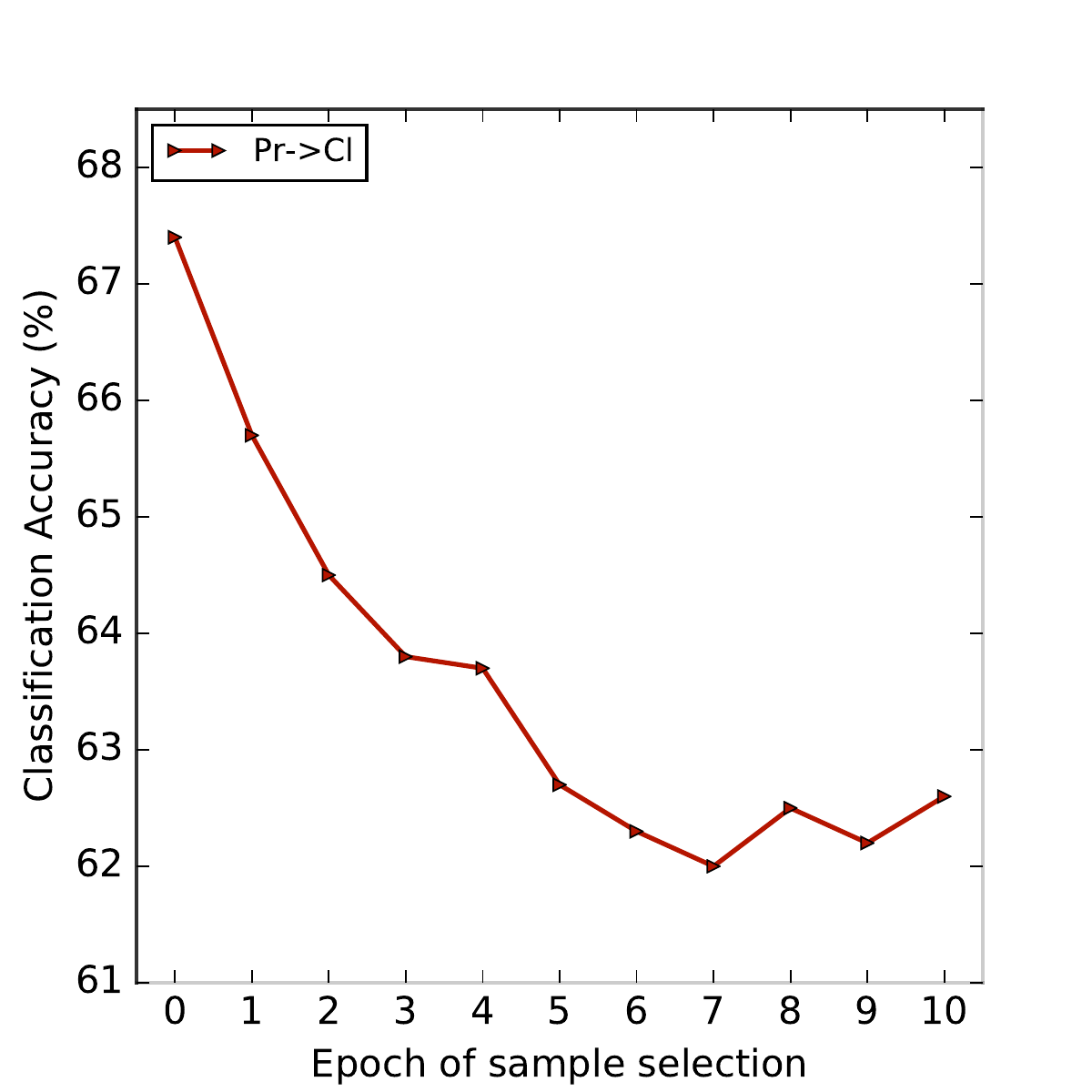}
\subcaption{Pr$\rightarrow$Cl}
\end{minipage}
\begin{minipage}[b]{.5\linewidth}
\centering
\includegraphics[height=2.8cm,width=5.5cm]{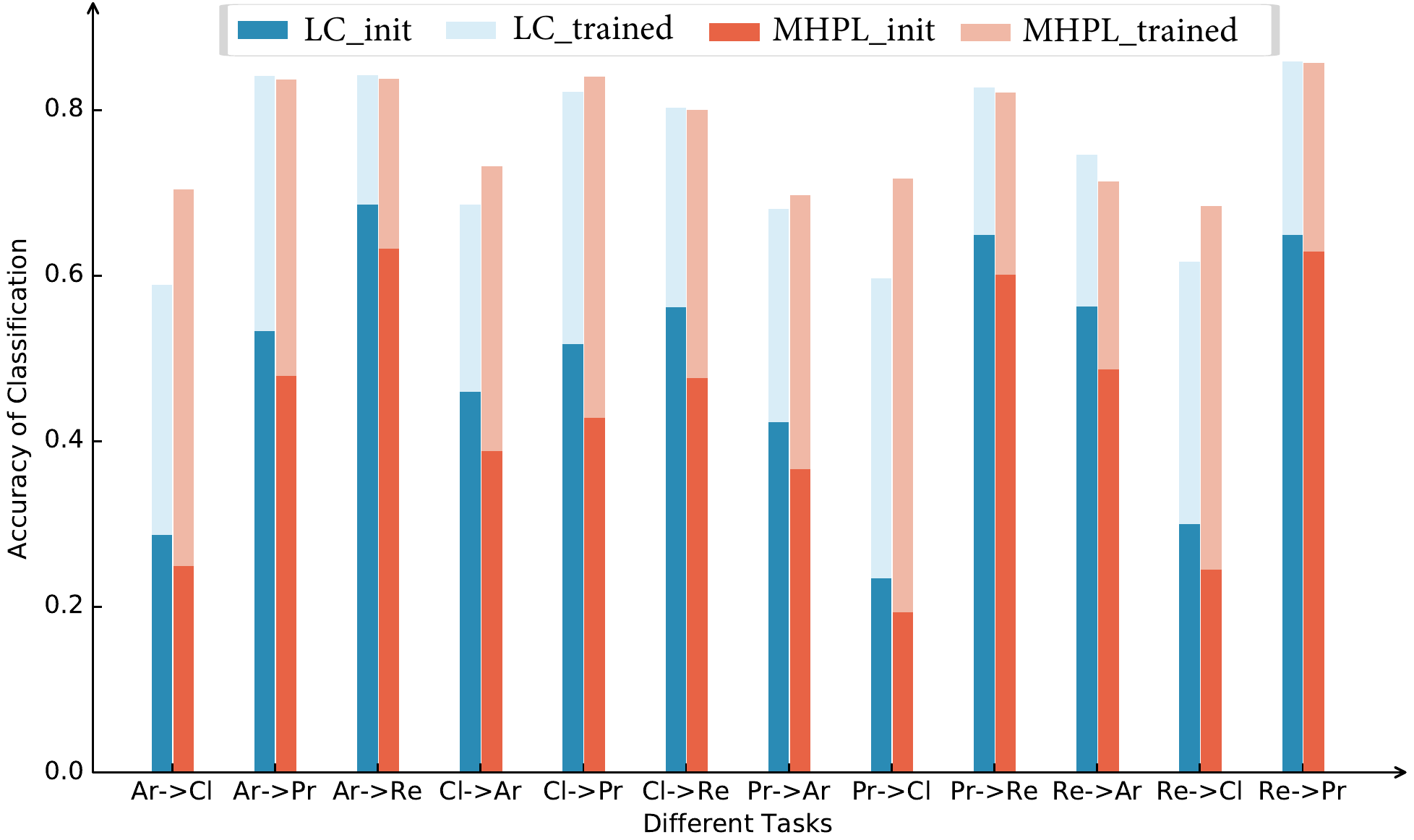}
\subcaption{Accuracy on neighbors}
\end{minipage}
\caption{Ablation studies on the selection criterion.  (a) represents the accuracy of key components from NAU and NDR at different selection ratios. (b) and (c) represent the feature visualization with and without NDR, respectively. They demonstrate the consistent advantages of NDR on representation learning as Fig. \ref{Fig_result_feature_visualization}. (d) and (e) show the model benefits brought by samples selected at different epochs. (f) shows the accuracy of  initial neighbors before and after training.}
\label{Fig_key_component}
\end{figure}

\noindent\textbf{Ablation on neighbor ambient uncertainty and neighbor diversity relaxation (NDR).} To investigate the efficacy of key components of our criterion for selecting minimum happy points, we firstly conduct an ablation study with the following variants on Ar$\rightarrow$Cl in various sample selection ratios from 1\% to 10\%: (i) MHPL w/o neighbor purity; (ii) MHPL w/o neighbor affinity; (iii) MHPL w/o NDR; (iv) MHPL. As shown in Fig.  \ref{Fig_key_component} (a), the full method outperforms other variants in various selection ratios, indicating the necessity of each key component. In addition, as shown in Fig. \ref{Fig_key_component} (b) and (c), the samples selected with NDR have more diversity among the high overlapping regions.

\noindent\textbf{Ablation on one-shot querying.} 
In order to verify the effectiveness of one-shot querying, we compare the effect with samples selected at different epochs of model training. As shown in Fig. \ref{Fig_key_component} (d) and (e), the abscissa indicates the epoch of model training for sample selection, where epoch zero represents the source model. It is observed that the samples selected by the source model obtain larger model performance gains on Cl$\rightarrow$Pr and Pr$\rightarrow$Cl.
Additionally, when sample selection is done on larger epochs, the model performance is significantly degraded, as the trained model cannot properly reflect the distribution discrepancy, resulting in the difficulty to explore the target-like samples.

\noindent\textbf{Effect on correcting neighbors.} 
We further analyze the role of selected samples in correcting the pseudo-labels of their neighbors and compare MHPL with the state-of-the-art LC \cite{he2019towards}. Fig. \ref{Fig_key_component} (f) shows the accuracy of the selected sample's initial neighbors before (dark color) and after training (light color). The lighter the color, the better the performance. It is obvious that MHPL is better at correcting neighbors than LC in all tasks, especially in challenging tasks Ar$\rightarrow$Cl, Pr$\rightarrow$Cl, and Re$\rightarrow$Cl. After training, the accuracy of initial neighbors improves more significantly than it did before training, demonstrating that selected samples have better guidance on their confusing neighbors. 

\begin{figure}
\begin{minipage}[b]{.24\linewidth}
\centering
\includegraphics[height=2.8cm,width=3.1cm]{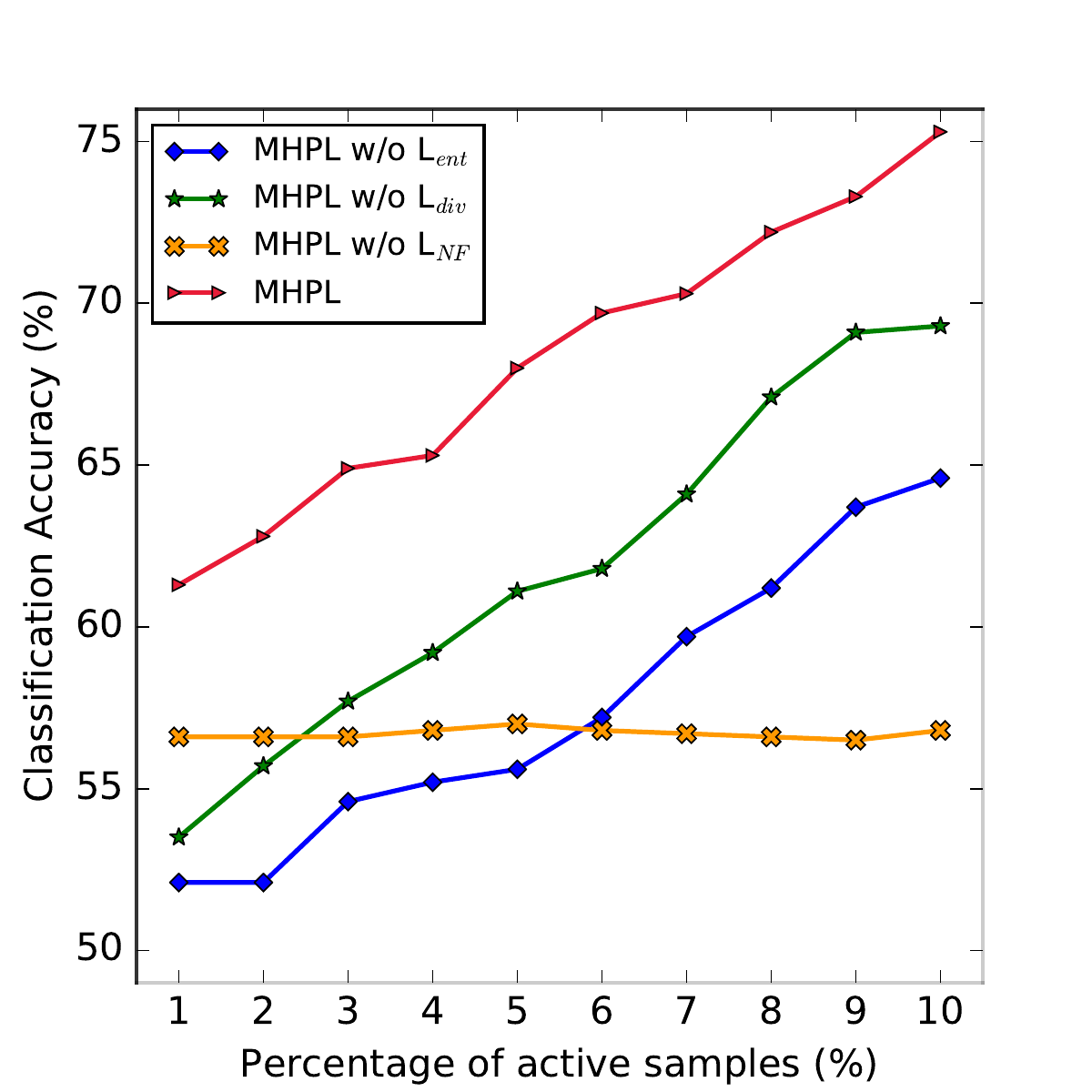}
\subcaption{Ar$\rightarrow$Cl}
\end{minipage}
\begin{minipage}[b]{.24\linewidth}
\centering
\includegraphics[height=2.8cm,width=3.1cm]{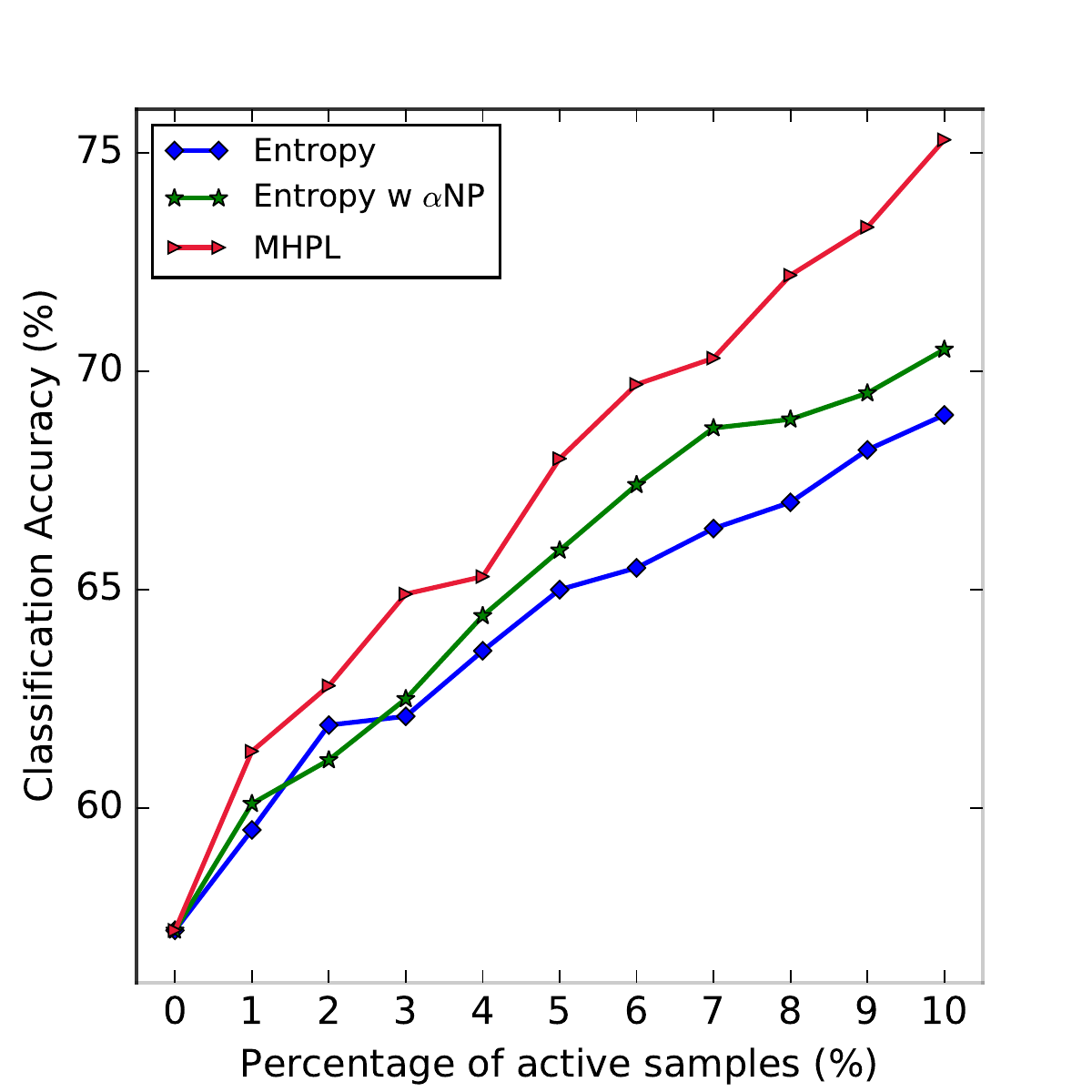}
\subcaption{Ar$\rightarrow$Cl}
\end{minipage}
\begin{minipage}[b]{.24\linewidth}
\centering
\includegraphics[height=2.8cm,width=3.1cm]{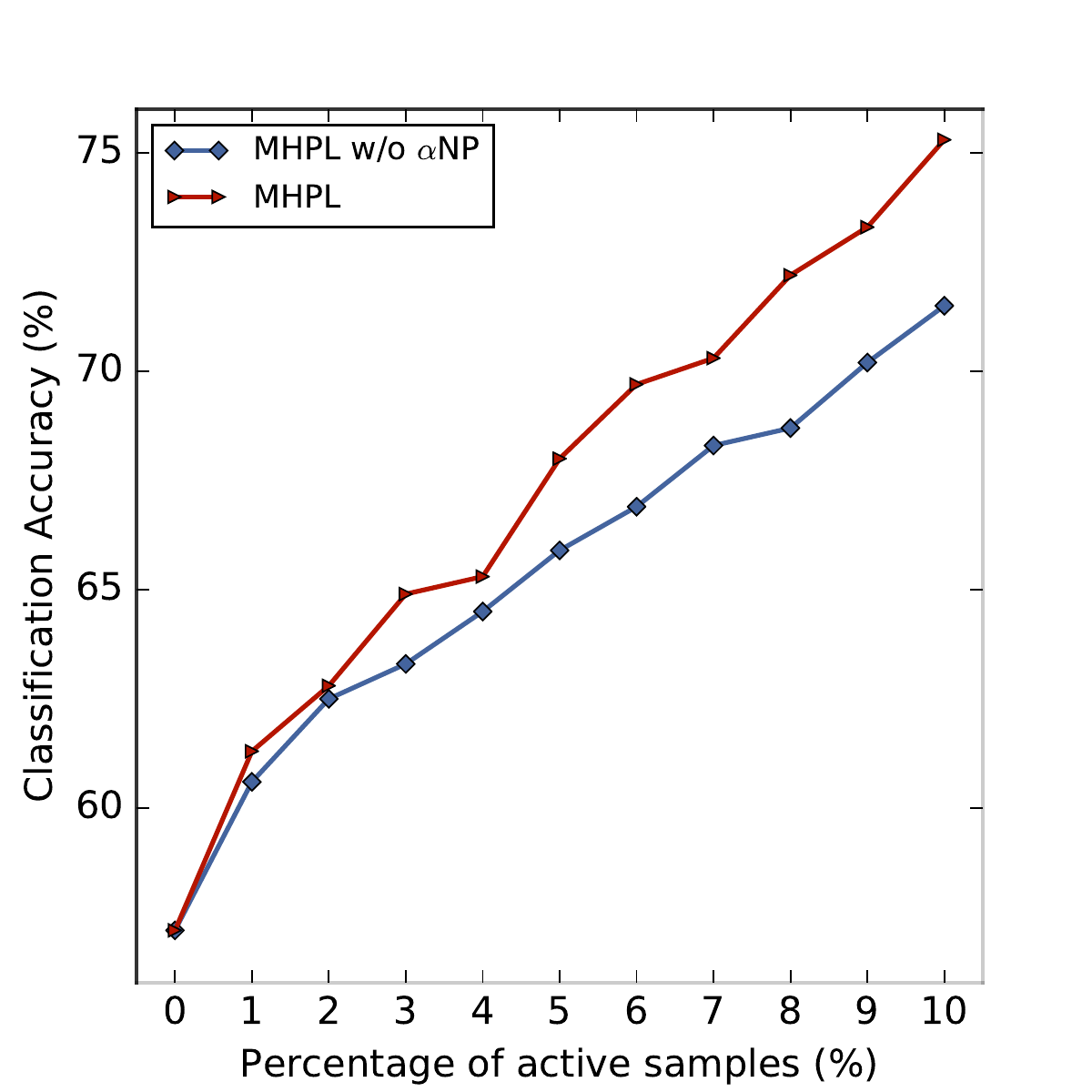}
\subcaption{Ar$\rightarrow$Cl}
\end{minipage}
\begin{minipage}[b]{.24\linewidth}
\centering
\includegraphics[height=2.8cm,width=3.1cm]{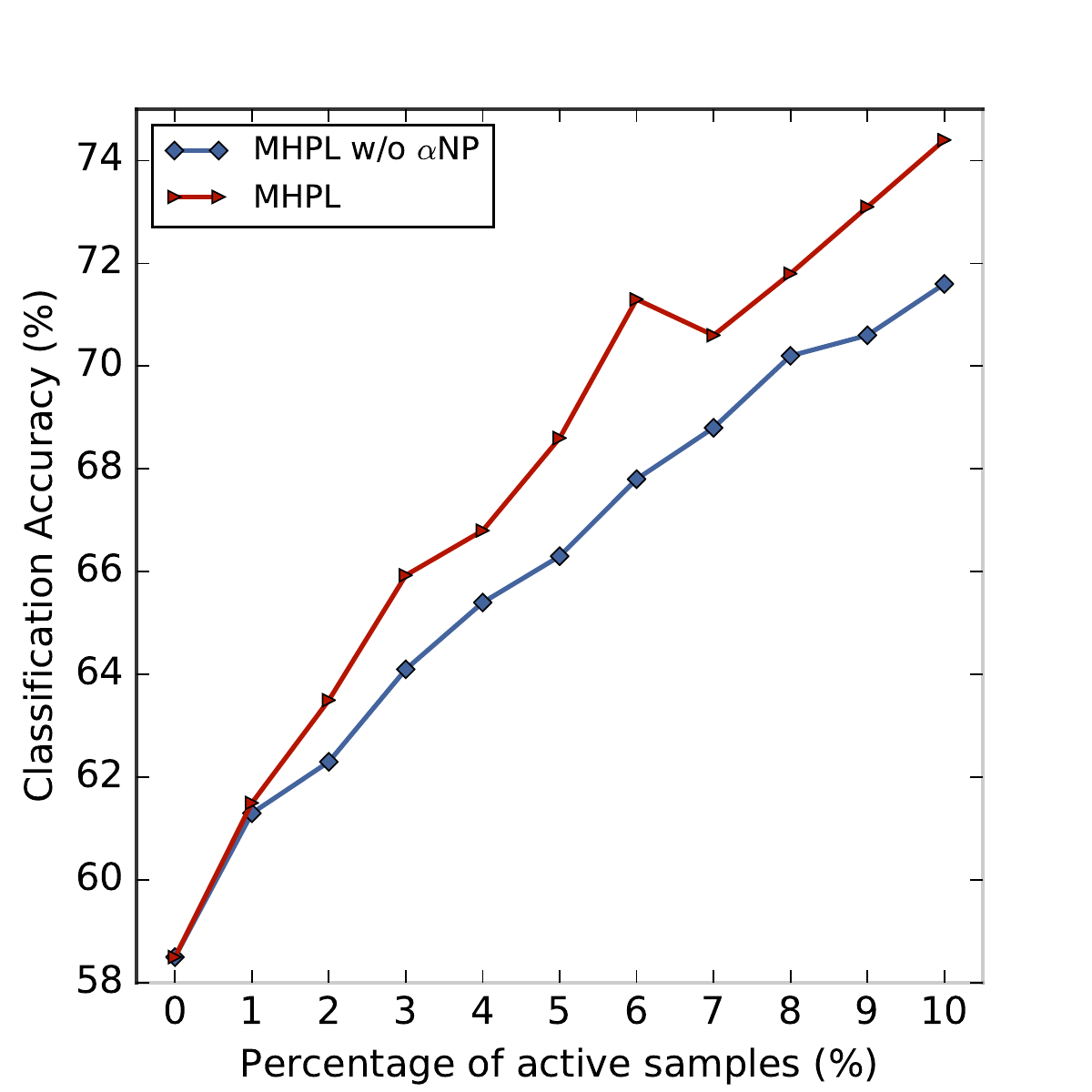}
\subcaption{Re$\rightarrow$Cl}
\end{minipage}
\caption{
Ablation studies on loss functions.}
\label{Fig_result_ablation}
\end{figure}

\noindent\textbf{Ablation on loss functions.} To demonstrate the effect of loss functions in Eq. \ref{Eq12:Totalloss}, we perform ablation studies on Ar$\rightarrow$Cl at various sample selection ratios. As shown in Fig. \ref{Fig_result_ablation} (a), the effect of NF loss increases as more active samples are chosen. Meanwhile, entropy loss and KL divergence also promote model learning. After removing NF loss (yellow), the accuracy of different selection ratios remains the same because the selected samples do not participate in training and the model is always trained using entropy loss and KL divergence. To verify the versatility of NF loss, we integrate it into other sample selection strategies. Specially, we assign the weight, $\alpha\text{NP}$, to the samples selected by entropy, and the value of $\text{NP}$ is calculated by entropy \cite{DBLP:conf/ijcnn/WangS14}. As shown in Fig. \ref{Fig_result_ablation} (b), the accuracy with $\alpha\text{NP}$ (green) is always better than entropy with standard cross-entropy loss (blue) on Ar$\rightarrow$Cl, while their results are still lower than our MHPL (red). Moreover, after removing the weight $\alpha\text{NP}$ of NF Loss on Ar$\rightarrow$Cl and Re$\rightarrow$Cl, the accuracy becomes lower, as shown in Fig. \ref{Fig_result_ablation} (c) and (d).

\section{Related Work}
\noindent\textbf{Domain Adaptation} (DA)
 aims to transfer the knowledge from amounts of labeled source data to the unlabeled target domain. Most DA works attempt to align the feature distributions across domains with moment matching \cite{long2015learning,MMD,long2017deep} or adversarial learning \cite{ganin2015unsupervised,saito2018maximum,long2018conditional}. Recently, semi-supervised DA \cite{DBLP:conf/iccv/SaitoKSDS19,kim2020attract,jiang2020bidirectional} and few-shot DA \cite{teshima2020few} have verified that utilizing a few labeled target data is helpful to performance benefits. However, the success of the above methods depends on amounts of annotated source data, which is unrealistic and unpractical  in privacy-persevering scenarios.

\noindent\textbf{Source Free  Domain Adaptation} (SFDA) is considered in the literature recently for relaxing the dependence on the source data. Existing SFDA works have made a great effort based on two categories of methods: \emph{i.e.,} source model exploration and data generation, while they face the  \emph{effect bottleneck} with limited performance gains.
The works based on source model exploration, SHOT \cite{liangjian}, HCL \cite{huang2021model}, NRC \cite{yang2021exploiting}, and A$^2$Net \cite{xia2021adaptive}, aim to utilize the potential distribution information stored in the given model. However, these methods cannot guarantee the positive transfer with limited information, without more than 1\% improvement over the initial SHOT \cite{liangjian} on Office-Home \cite{venkateswara2017deep}. Another line of work based on data generation,  MA \cite{li2020model} and CPGA \cite{qiu2021source}, intends to generate the source or target data to obtain supervised information, costing amounts of time and resources. But these works also perform worse in practice as the real samples or correct feature prototypes are difficult to generate. Hypothesis Transfer Learning (HTL) \cite{kuzborskij2013stability,ahmed2020camera} aims to transfer the knowledge using learned source hypotheses and limited labeled target data without the source data. However, naively selecting labeled target data would be inefficient. In this paper, we introduce the active source free domain adaptation, aiming to break through the  \emph{effect bottleneck}
with a few actively labeled target data.

\noindent\textbf{Active Learning} (AL) aims to obtain a satisfactory model at the cost of a limited annotation budget. Existing AL methods are mainly based on two categories: (1) Uncertainty, \emph{e.g.,} entropy \cite{DBLP:conf/ijcnn/WangS14} and BVSB \cite{DBLP:conf/cvpr/JoshiPP09}; (2) Diversity, \emph{e.g.,} CoreSet \cite{DBLP:conf/iclr/SenerS18}. However, neither method is suitable for domain adaptation under domain shift.
Recently, Active DA introduces active learning to domain adaptation. AADA \cite{DBLP:conf/wacv/SuTSLMC20} leverages the domain discriminator to qualify the target-like samples. TQS \cite{DBLP:conf/cvpr/FuC0L21} utilizes ensemble learning to select informative samples by transferable committee, uncertainty, and domainness. Clue \cite{DBLP:conf/iccv/PrabhuCSH21}  combines uncertainty and diversity to select samples. EADA \cite{xie2021active} determines the key samples with free energy \cite{liu2020energy}. Due to domain shift and unavailable source data, it is impossible to directly apply the existing criteria of AL and Active DA to ASFDA. 

\section{Conclusion}
In this paper, we propose a new setting, active source free domain adaptation (ASFDA), for the first time. ASFDA maximizes the model benefits with the minimum cost of data annotation under domain shift in the privacy-preserving scenarios. We firstly find the essential minimum happy (MH) points that satisfy the properties of neighbor-chaotic, individual-different, and target-like. We then design the minimum happy points learning to explore and exploit the MH points well. Extensive experiments verify that we lay a promising baseline of ASFDA for further works.

{
\small
\bibliographystyle{unsrt}
\bibliography{neurips_2022}
}

\section*{Checklist}

\begin{enumerate}

\item For all authors...
\begin{enumerate}
  \item Do the main claims made in the abstract and introduction accurately reflect the paper's contributions and scope?
    \answerYes{}
  \item Did you describe the limitations of your work?
    \answerYes{The limitations is attached in Appendix A.}
  \item Did you discuss any potential negative societal impacts of your work?
    \answerYes{The potential negative social impacts is attached in Appendix B.}
  \item Have you read the ethics review guidelines and ensured that your paper conforms to them?
    \answerYes{}
\end{enumerate}

\item If you are including theoretical results...
\begin{enumerate}
  \item Did you state the full set of assumptions of all theoretical results?
    \answerNA{}
        \item Did you include complete proofs of all theoretical results?
    \answerNA{}
\end{enumerate}

\item If you ran experiments...
\begin{enumerate}
  \item Did you include the code, data, and instructions needed to reproduce the main experimental results (either in the supplemental material or as a URL)?
    \answerYes{We attach the code in the supplemental material.}
  \item Did you specify all the training details (e.g., data splits, hyperparameters, how they were chosen)?
    \answerYes{Detail implementation details is attached in Appendix C.}
        \item Did you report error bars (e.g., with respect to the random seed after running experiments multiple times)?
    \answerYes{All main results are average over three running times in all experiments}
        \item Did you include the total amount of compute and the type of resources used (e.g., type of GPUs, internal cluster, or cloud provider)?
    \answerYes{Detail information is attached in Appendix C.}
\end{enumerate}

\item If you are using existing assets (e.g., code, data, models) or curating/releasing new assets...
\begin{enumerate}
  \item If your work uses existing assets, did you cite the creators?
    \answerYes{}
  \item Did you mention the license of the assets?
    \answerNo{}
  \item Did you include any new assets either in the supplemental material or as a URL?
    \answerNo{}
  \item Did you discuss whether and how consent was obtained from people whose data you're using/curating?
    \answerNo{}{}
  \item Did you discuss whether the data you are using/curating contains personally identifiable information or offensive content?
    \answerNo{}
\end{enumerate}

\item If you used crowdsourcing or conducted research with human subjects...
\begin{enumerate}
  \item Did you include the full text of instructions given to participants and screenshots, if applicable?
    \answerNA{}
  \item Did you describe any potential participant risks, with links to Institutional Review Board (IRB) approvals, if applicable?
    \answerNA{}
  \item Did you include the estimated hourly wage paid to participants and the total amount spent on participant compensation?
    \answerNA{}
\end{enumerate}

\end{enumerate}

\end{document}